\documentclass[oribibl]{llncs}
\usepackage[numbers]{natbib}
\usepackage{url}

\usepackage{graphicx}
\usepackage{amssymb}
\usepackage{amsmath, xparse}
\usepackage{subcaption}
\usepackage[export]{adjustbox}
\usepackage[hidelinks]{hyperref}
\usepackage[dvipsnames]{xcolor}
\begin{document}
\title{{\scriptsize Author's accepted manuscript of:  Velarde, G. (2022). Forecasting with Deep Learning [White Paper]. Vodafone. \textit{The Data Digest, 2(8)}.} \\ \textcolor{white}{.}
 \\
Forecasting with Deep Learning}
%
%
\author{Gissel Velarde, Ph.D.} 
%
%
\institute{Data \& Analytics, Vodafone \\
\email{gissel.velarde@vodafone.com}} 
%
\maketitle              
\begin{abstract}
This paper presents a method for time series forecasting with deep learning and its assessment on two datasets. The method starts with data preparation, followed by model training and evaluation. The final step is a visual inspection. Experimental work demonstrates that a single time series can be used to train deep learning networks if time series in a dataset contain patterns that repeat even with a certain variation. However, for less structured time series such as stock market closing prices, the networks perform just like a baseline that repeats the last observed value. The implementation of the method as well as the experiments are open-source. 

\end{abstract}
\section{Introduction}
This paper aims to present a method based on two related deep learning architectures: Long-Short Term Memory (LSTM) and Gated Recurrent Unit (GRU). Deep learning networks of the type Recurrent Neural Networks (RNNs) are known to model dependencies over time \cite{rumelhart1986learning}. Therefore, they are relevant in time series forecasting. LSTM decides to keep content thanks to its input, forget and output gates \cite{hochreiter1997long}. GRU consists of reset and update gates \cite{cho2014properties}. Since their inception, both networks have been extensively used in problems of sequential nature. 

Although there are classical methods for time series forecasting, such as Autoregressive Integrated Moving Average (ARIMA), this report focuses on exploring deep learning networks. Indeed, previous studies have shown that LSTM outperforms ARIMA on financial data \cite{siami2018comparison} and from various deep learning models, LSTM and GRU deliver low forecasting error \cite{balaji2018applicability}. Next, the method is explained in a nutshell. Its detailed description can be found in \cite{velarde2022open}.

\begin{figure}[h] 
\caption{A visual summary of the method.}\label{F1}
\includegraphics[width=1\textwidth]{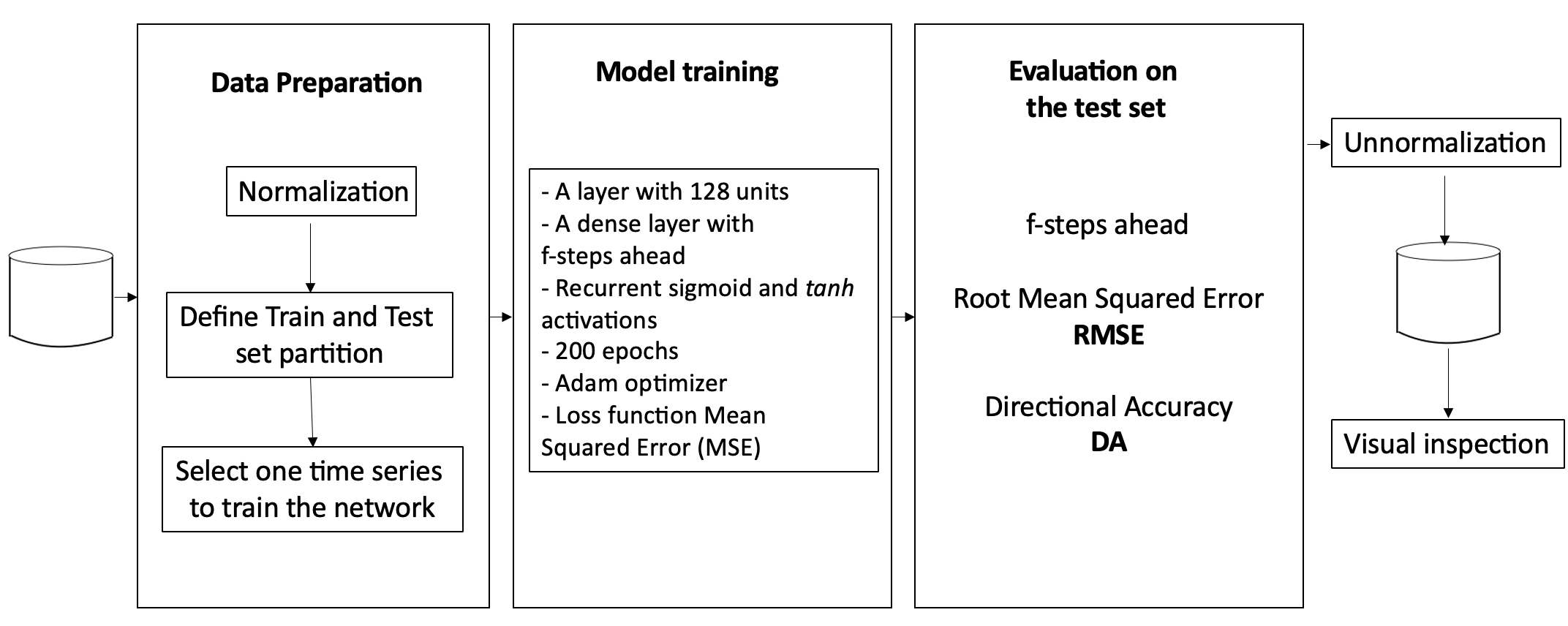}
\end{figure}

\section{Method}
The method consists of data preparation, model training, evaluation, and visual inspection, as seen in Fig. \ref{F1}. \textbf{Data Preparation} consists of normalization, the definition of train and test sets partition, and the selection of a time series for training. First, each time series in the data set is normalized between 0 and 1. Then, a time series of length $Q$ samples is prepared as in Fig. \ref{F1}, where $w$ is the window size, $f$ is the number of steps ahead for forecasting, and $N$ is the number of training samples. The remaining samples are used for testing. 

\begin{figure}[h] 
\caption{Data partitioning. From \cite{velarde2022open} }\label{F1}
\includegraphics[width=1\textwidth]{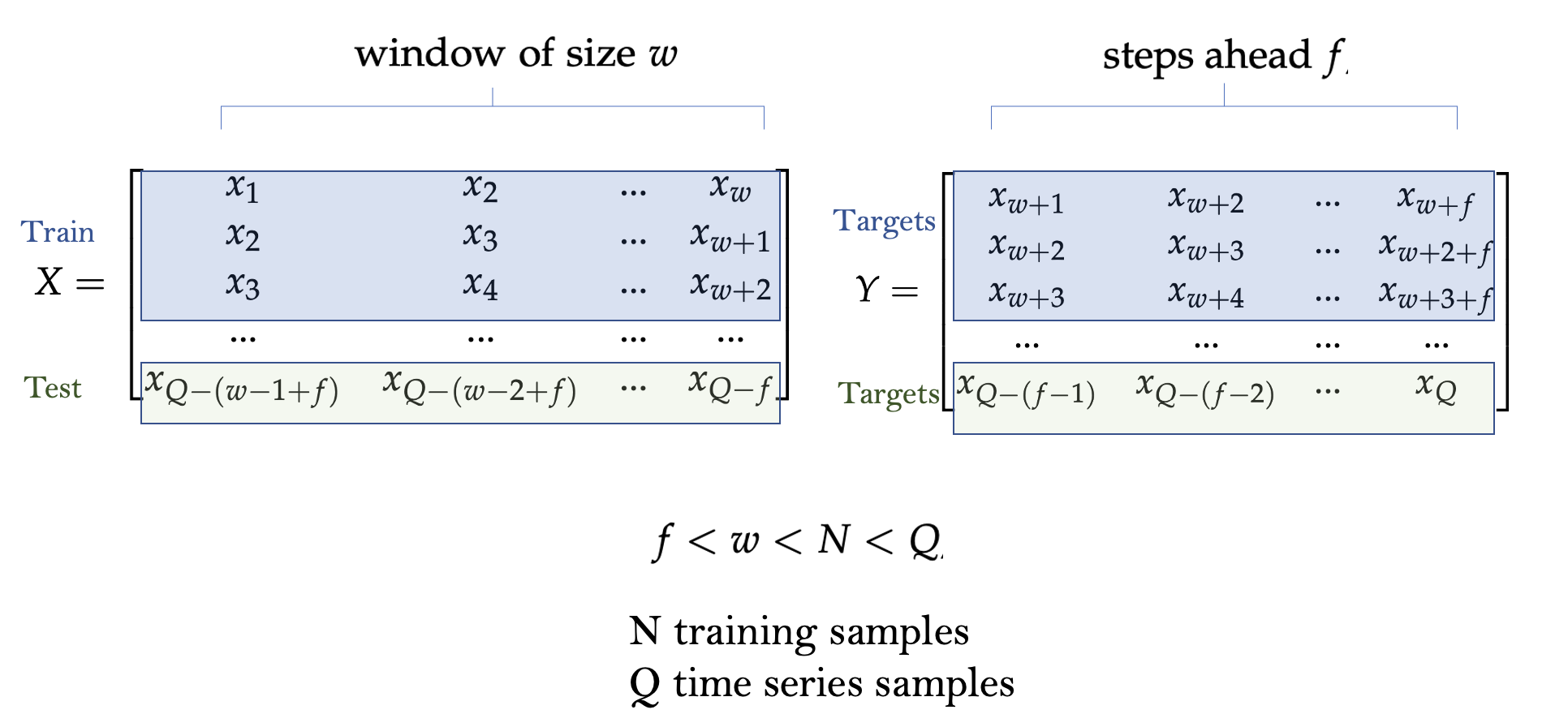}
\end{figure}

\textbf{Model training} consists of training either an LSTM or a GRU network with a layer of 128 units, followed by a dense layer that outputs $f$-step ahead. The networks are trained for 200 epochs, with Adam optimizer, and Mean Squared Error (MSE) Loss function.

\textbf{Evaluation} consists of measuring Root Mean Squared Error (RMSE) and Directional Accuracy (DA) between actual and predicted values on the test set. Finally, each time series is unnormalized and plotted for visual inspection to better understand the results. 

\begin{figure}[h]
\caption{Experimental setup.}\label{F3}
\includegraphics[width=1\textwidth]{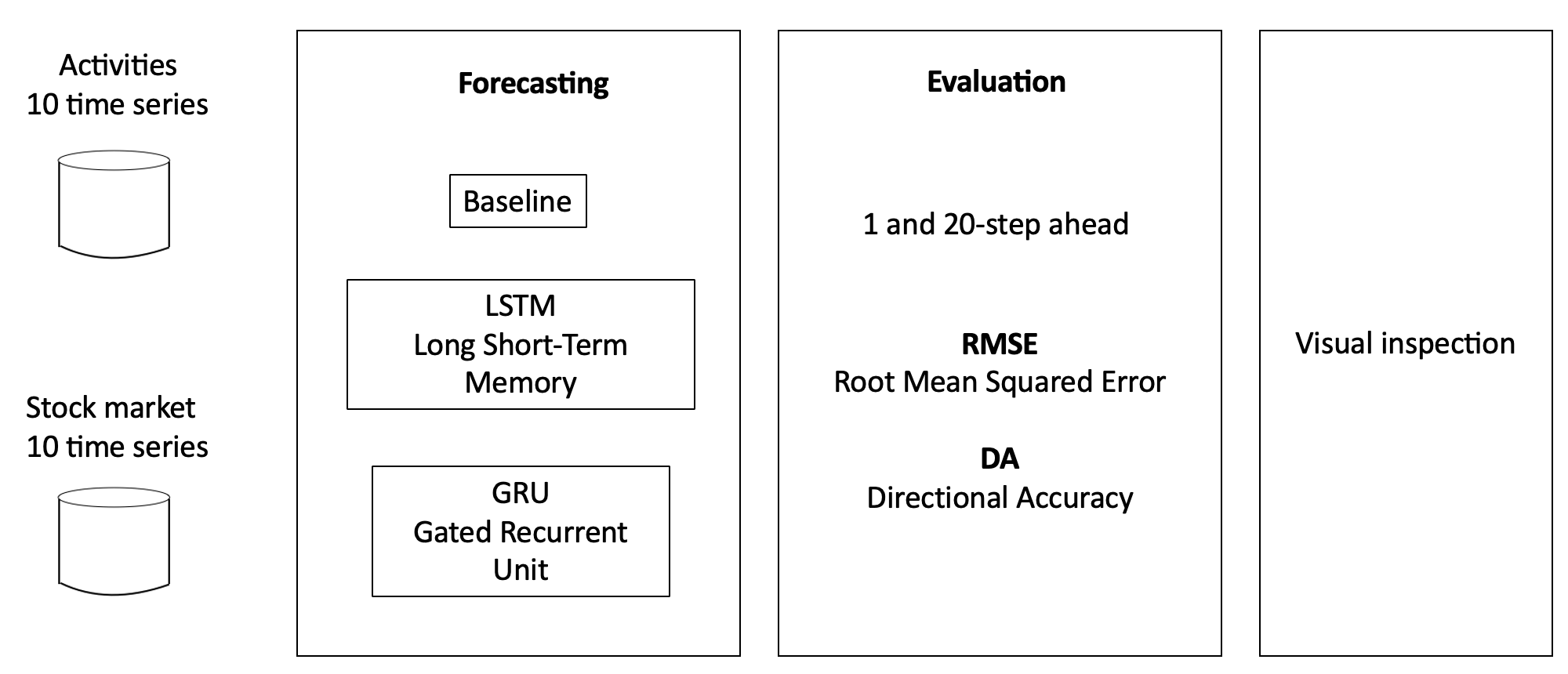}
\end{figure}

\begin{figure}[]
\caption{Activities dataset (synthetic). The first 100 samples are plotted. The dataset contains 3 584 samples per series. Taken from \cite{velarde2022open}.
} \label{FA}
\includegraphics[width=0.7\textwidth]{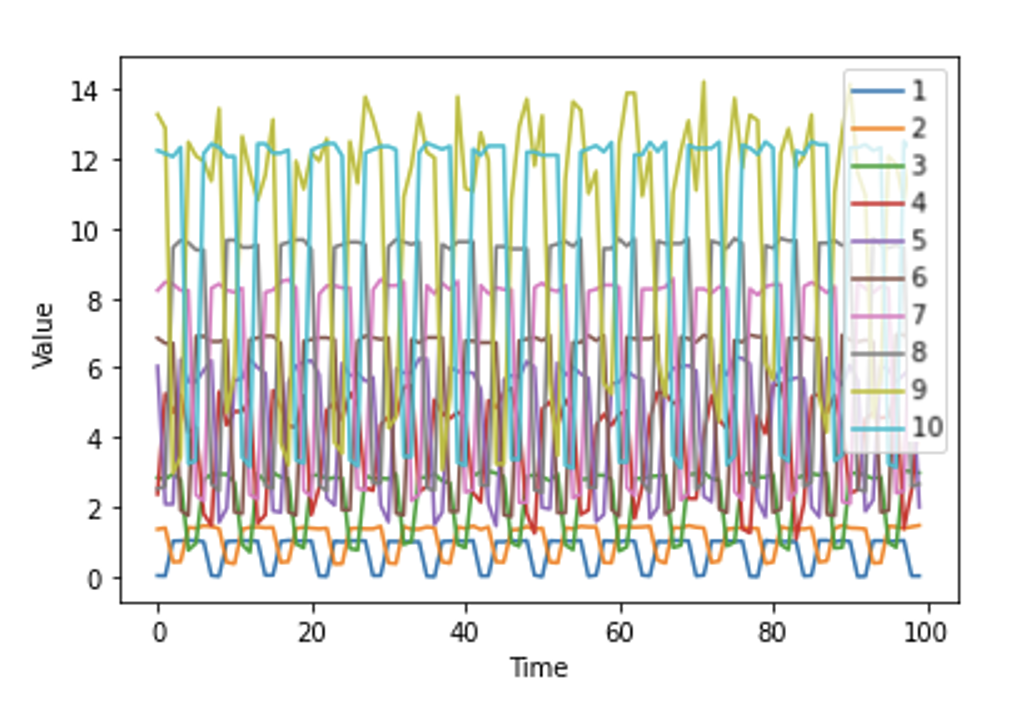}
\end{figure}

\begin{figure}[]
\caption{BANKEX Dataset. Closing Price in Indian Rupee (INR). 
Daily samples retrieved between 12 July  2005 and 3 November 2017.
All time series with 3 032 samples.  Taken from \cite{velarde2022open}.
} \label{FB}
\includegraphics[width=1\textwidth]{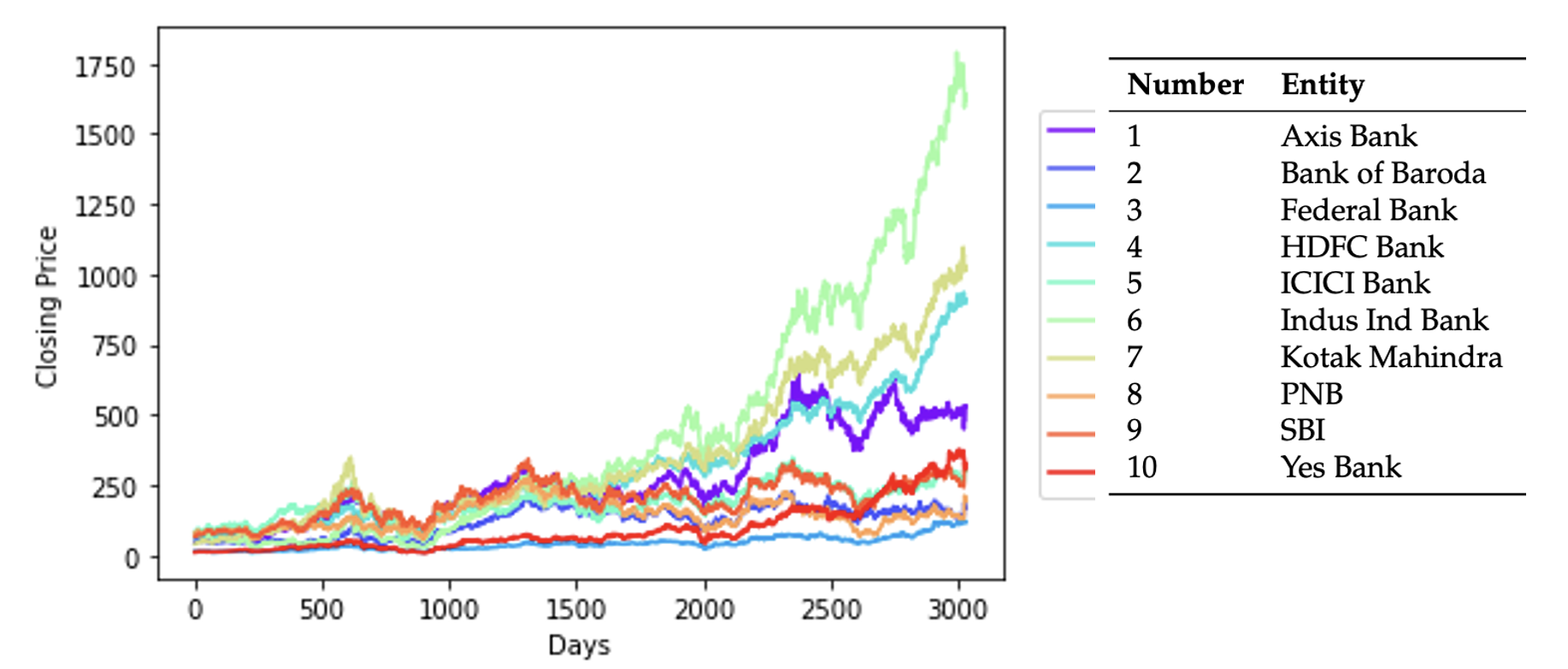}
\end{figure}

\begin{figure}[]
\caption{Normalized BANKEX Dataset. Daily samples retrieved between 12 July  2005 and 3 November 2017
All time series with 3 032 samples.  Taken from \cite{velarde2022open}.
} \label{FB01}
\includegraphics[width=1\textwidth]{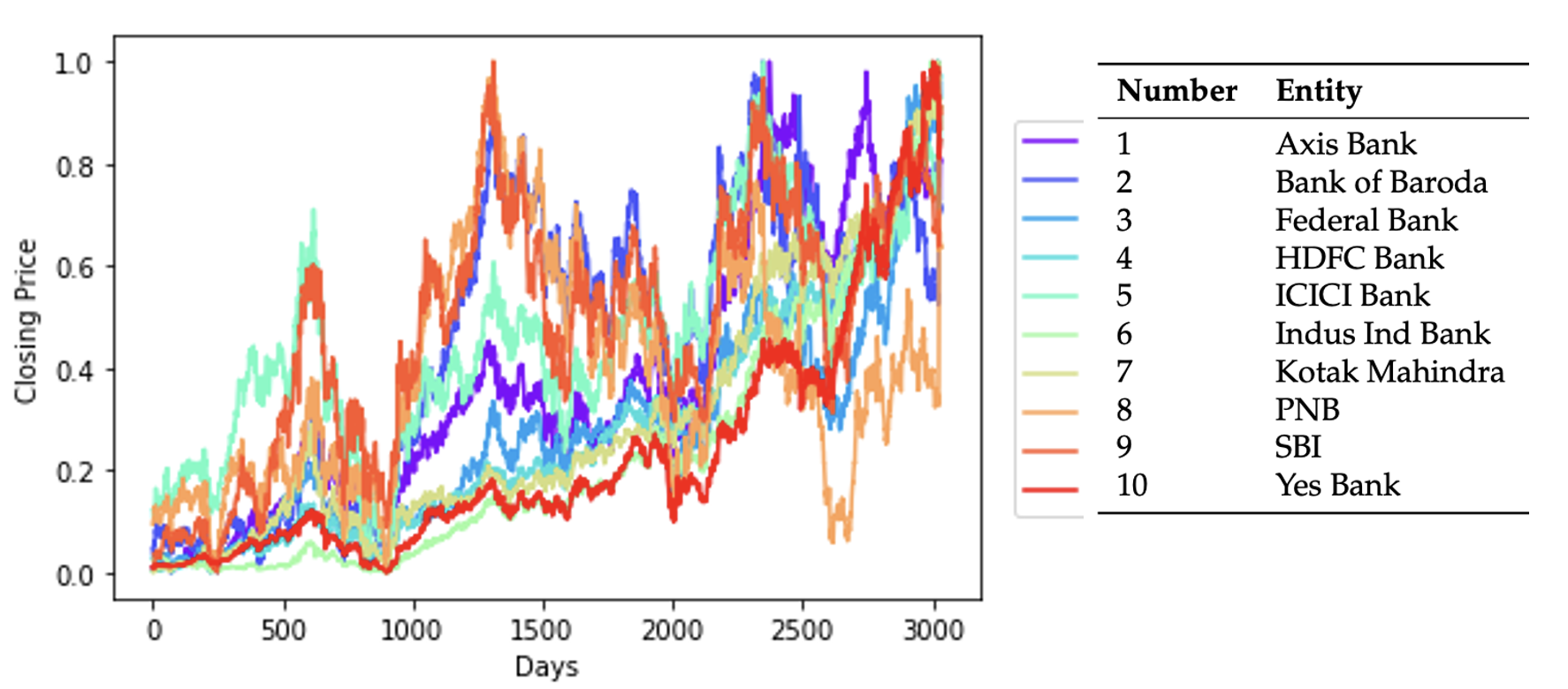}
\end{figure}

\section{Experiments}
The experimental setup can be seen in Fig. \ref{F3}. The method has been tested on two datasets, each with ten time series. The first dataset is the Activities dataset, which contains ten synthetic time series with five days of high activity and two days of low activity. This dataset may resemble, for example, the volume of weekly calls, see Fig. \ref{FA}. The second dataset is the BANKEX dataset, which contains stock market closing prices of ten financial institutions, see Fig. \ref{FB}. Fig. \ref{FB01} shows the effect of normalization between 0 and 1. A window of size $w$=60 days was used for data preparation with the first time series of each dataset. The last 251 samples of each series were used for testing. Forecasting was performed by LSTM, GRU networks, and a Baseline that simply repeats the last observed value. Each model was evaluated on one-step and twenty-step ahead RMSE and DA.  

\subsection{Results}
Tables 2 to 5 summarise the mean and standard deviation (SD) of RMSE and DA over the ten time series on the test set of each dataset. Close-to-zero RMSE and close-to-one DA are preferred. Tables 2 and 3 present the results on the Activities dataset. The best results are highlighted in blue.  

For One-Step ahead, GRU significantly outperforms LSTM and the Baseline on $RMSE$. However, both deep learning networks perform equally well on DA, and significantly outperform the Baseline. For Twenty-step ahead forecast, LSTM is the clear winner considering RMSE and DA. On the Activities dataset, the networks prove their capability to learn patterns that repeat, even with a certain variation.

\begin{figure}[]
\caption{An example of 1-step ahead forecast. Actual and predicted closing price over the first 100 days of the test set Yes Bank. Closing Price in Indian Rupee (INR). Taken from \cite{velarde2022open}.
}\label{FV}
\includegraphics[width=1\textwidth]{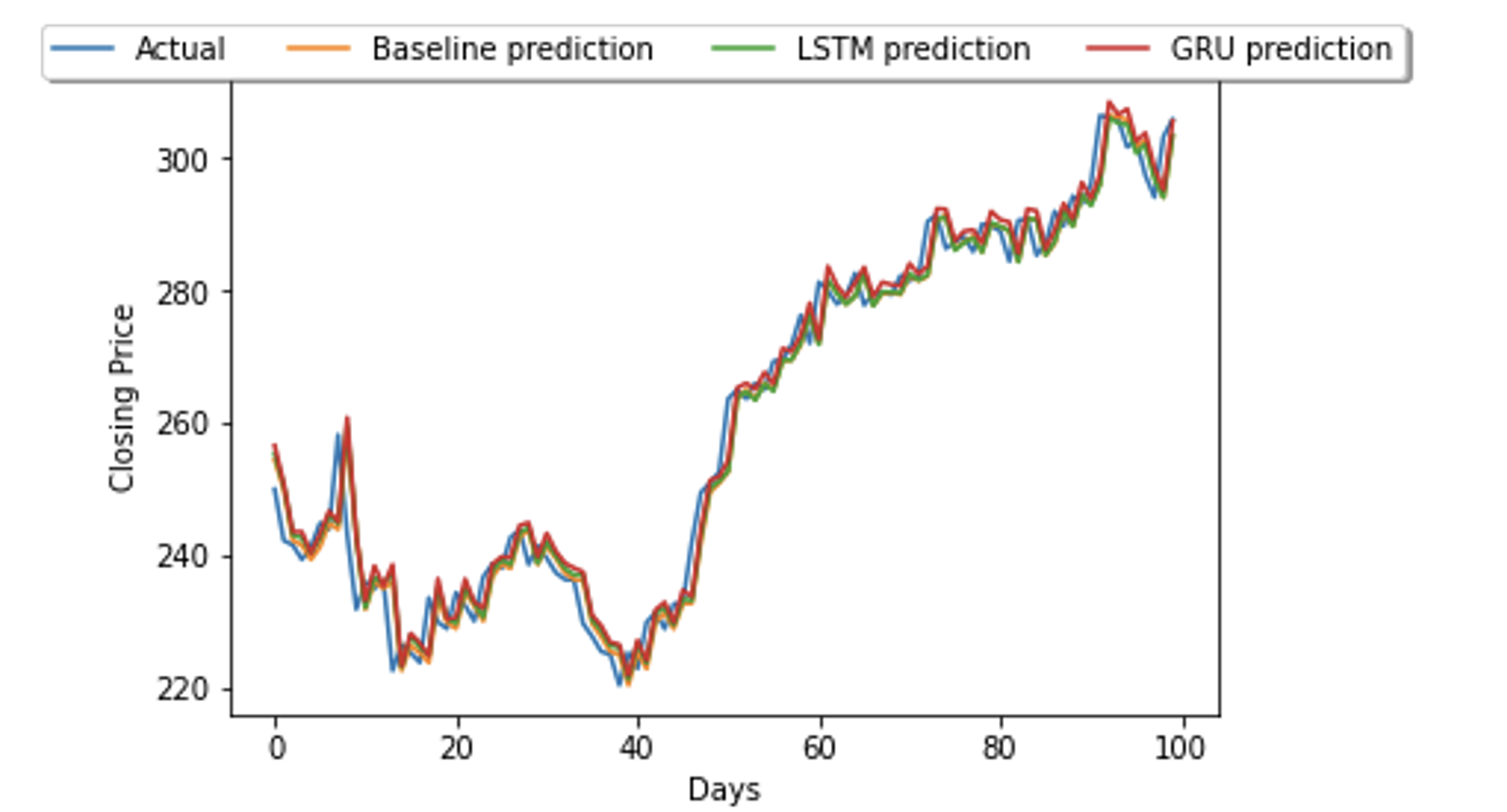}
\end{figure}

\begin{figure}[]
\caption{Examples of 20-step ahead forecast. Taken from \cite{velarde2022open}.
}\label{FV2}
\includegraphics[width=1\textwidth]{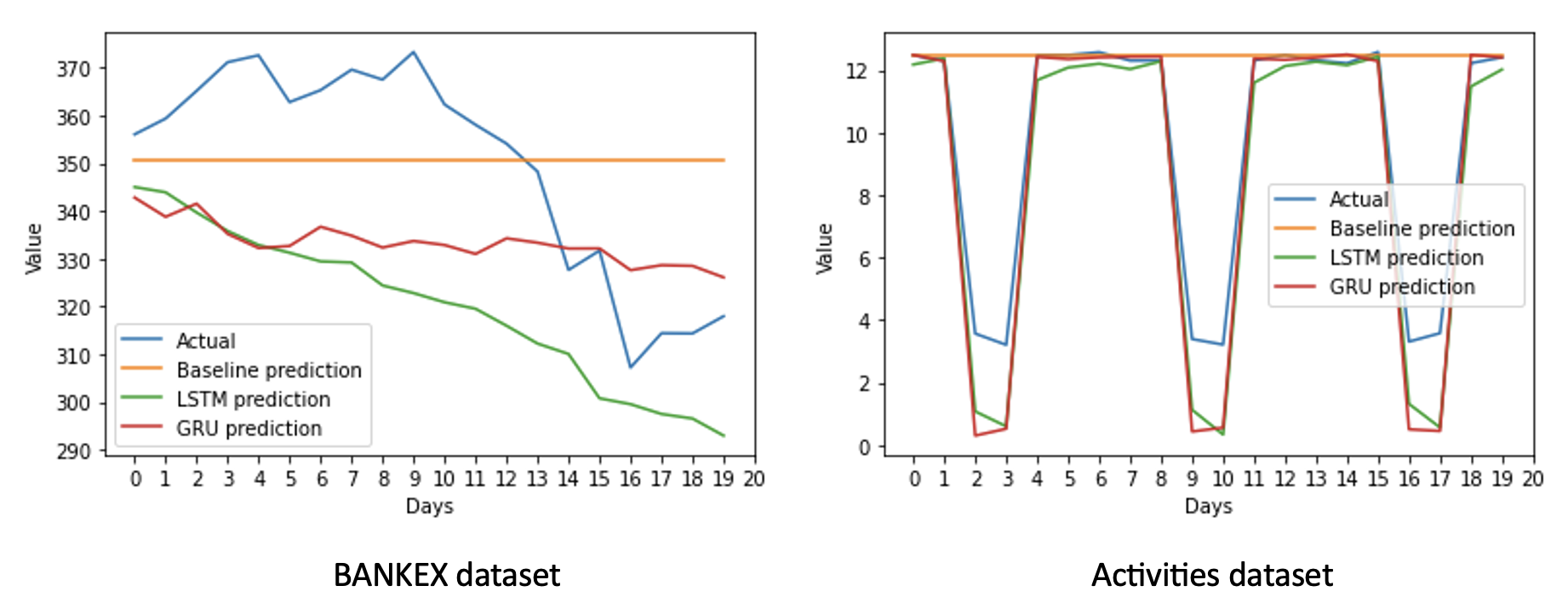}
\end{figure}

Tables 4 and 5 present the results on the BANKEX dataset. In this case, the networks perform just like the Baseline, possibly due to the nature of stock market series. Finally, visual inspection helps understand the numerical results; see Fig. \ref{FV} and Fig. \ref{FV2}.

\begin{table}[]
\includegraphics[width=0.8\textwidth]{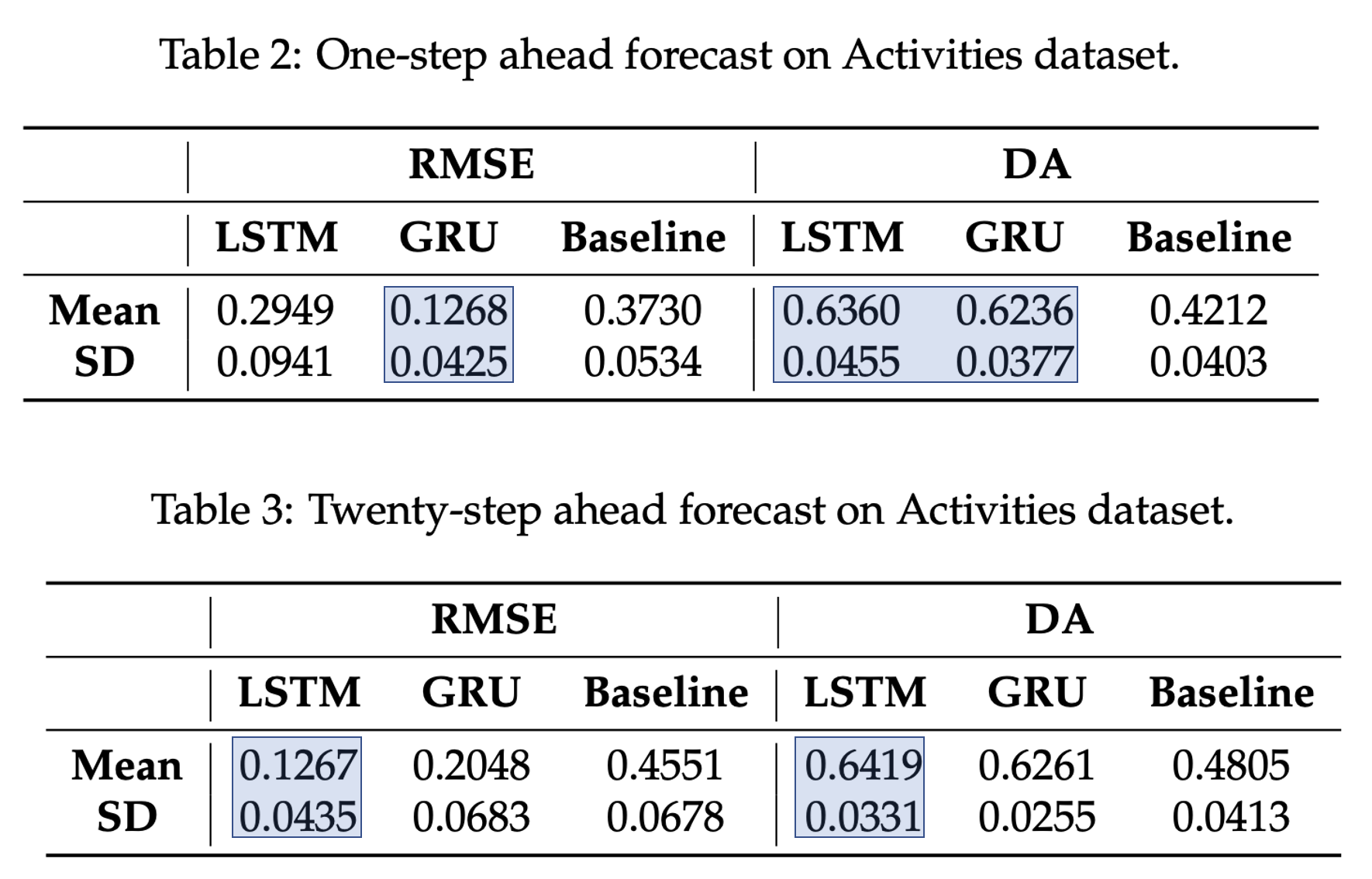}

\includegraphics[width=0.8\textwidth]{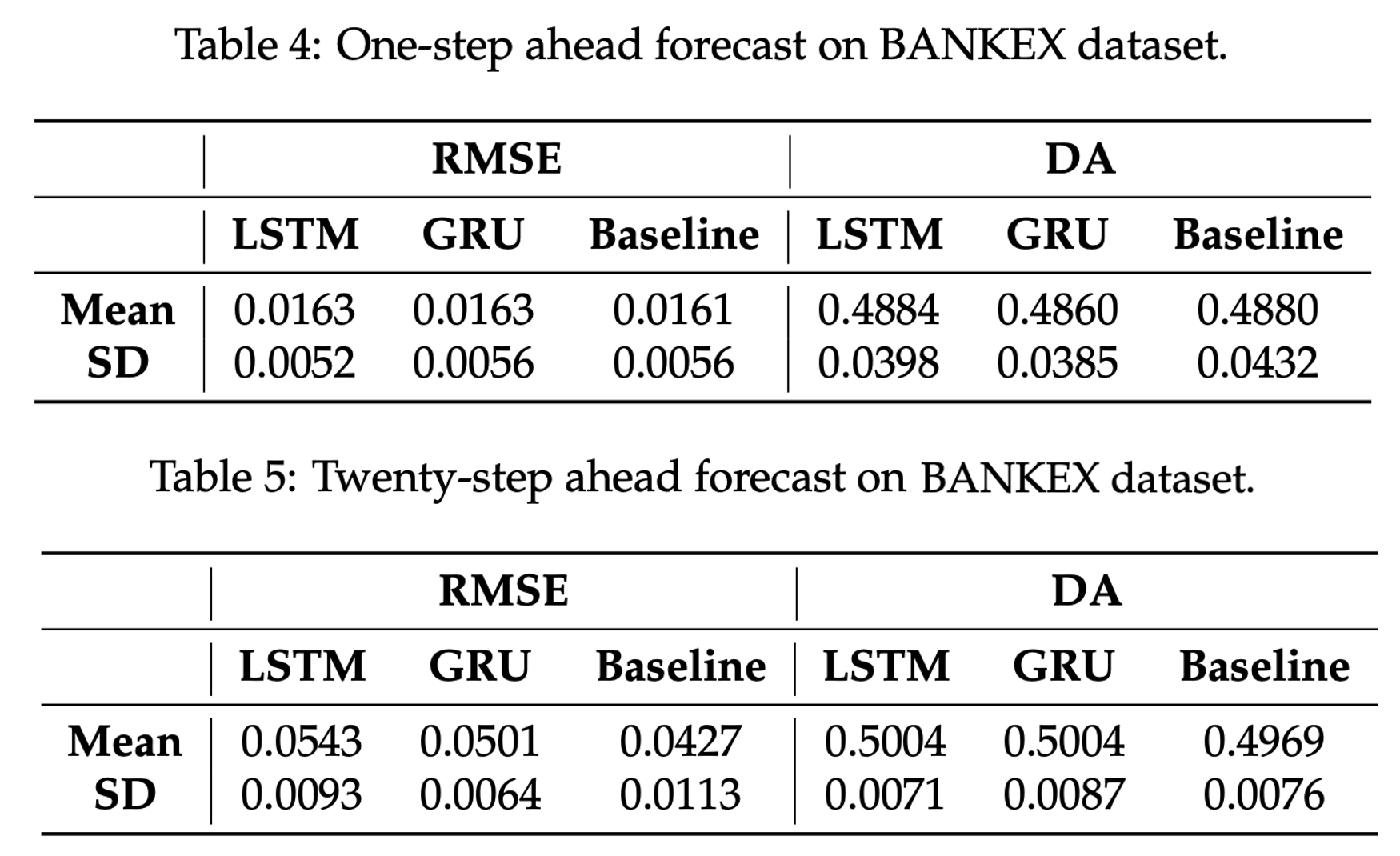}
\end{table}


\section{Conclusion}
This paper showcases a method using LSTM and GRU deep learning networks for time series forecasting with the following highlights:
\begin{itemize}
\item It shows that LSTM and GRU networks can be trained for forecasting with a single time series in a dataset of series with patterns that repeat even with certain variation, if the data is properly prepared.
\item It shows the performance of the method on two datasets. While the method is appropriate for time series that contain patterns that repeat like those of weekly activities, it is not appropriate for stock market data, possibly because some information is not encoded in closing price alone, or due to the problem's nature.
\item It is flexible to forecast not only one-step ahead but also twenty-step ahead.
\item In addition to the numerical evaluation provided by RMSE and DA, visual inspection helps understand the numerical results.
\item The implementation and results are reproducible and shared as open-source at: \href{https://github.com/Alebuenoaz/LSTM-and-GRU-Time-Series-Forecasting}{https://github.com/Alebuenoaz/LSTM-and-GRU-Time-Series-Forecasting}
\end{itemize}

\textcolor{white}{.}

\textbf{Dr. Gissel Velarde} is a Senior Expert Data Scientist at Vodafone. She holds a Ph.D. degree from Aalborg University for her thesis on Machine Learning-based methods for media analysis, pattern discovery, and classification. In addition, she developed computational creativity models. She taught Artificial Intelligence, Machine Learning, and Deep Learning courses at the university level. Besides, she supervised the development of analysis and recommendation systems for media applications. Currently, she leads projects for forecasting and fraud detection systems.

%
%
%
 \bibliographystyle{splncs04}
 \bibliography{References.bib}

\end{document}